# An Approach for Resource Sharing in Multilingual NLP

Manuela Kunze and Chun Xiao

*Otto-von-Guericke-Universität Magdeburg,*
Institut für Wissens- und Sprachverarbeitung,
P.O.box 4120,
D-39016 Magdeburg, Germany,
[makunze,xiao]@iws.cs.uni-magdeburg.de

**Abstract.** In this paper we describe an approach for the analysis of documents in German and English with a shared pool of resources. For the analysis of German documents we use a document suite, which supports the user in tasks like information retrieval and information extraction. The core of the document suite is based on our tool XDOC. Now we want to exploit these methods for the analysis of English documents as well. For this aim we need a multilingual presentation format of the resources. These resources must be transformed into an unified format, in which we can set additional information about linguistic characteristics of the language depending on the analyzed documents. In this paper we describe our approach for such an exchange model for multilingual resources based on XML.

Presently users must work with an increasing amount of documents. For a better efficiency of this work they need an effective system for information retrieval and information extraction. The tool XDOC[1] was designed and developed for information retrieval from documents and was first used with German documents from the casting domain and from forensic medicine. The document suite supports for example knowledge acquisition from technical documentation about casting technology [3] and analysis of autopsy protocols. XDOC contains several methods for linguistic and semantic analyses of German documents. The resources are implemented as lexica, collections of case frames and ontologies. Presently XDOCs results are complete syntactic parses of sentences and conceptual interpretations of syntactic structures. Now we work with a new pool of English documents from the domain of bioinformatics; we focus on bioinformation extraction from Medline abstracts. We try to extract information about enzymes and medical substances, as well as certain interactions between them, such as the inhibition relationship. For this project we want to reuse the methods of XDOC. Most methods of XDOC work independently from the specific language. Through the separate loading of resources we can use the methods for the analyses of documents in other languages, e.g. in English. For the description of the resources we use the XML standard. Through XML we get a descriptive format for our resources and they can be easily analyzed inside a test suite. The methods of XDOC and the required resources are depicted in figure 1. For structure detection we primarily need an abbreviation lexicon for the analyzed language. POS tagging is strongly dependent on the language, so we integrated an existing tagger (e.g. the Brill tagger [2]) for the English documents. For the next step we map the specific tagset of the used tagger to the tagset of the chart parser used in syntactic analysis. This parser also needs a grammar, which is dependent on the language or the specific characteristics of the language in the domain. The lexical categories of this grammar are given by the tagset of the chart parser. Semantic analyses work with different resources. The lexicon for the semantic tagger is based on the different wordclasses and can also be described with XML structures. A minor change must be done for the description of case frames. In the English language we must regard the positions of subject and object, which are not so relevant in German. The ontologies are the same, we only need the mapping of the linguistic characteristic of an entry into the ontology. The semantic structure analysis is based on the semantic analysis of syntactic structure, like e.g. structures of noun phrases. This can also be realized through a simple lexicon (see figure 2).

---

[1] XDOC stands for *X*ML based *doc*ument processing.

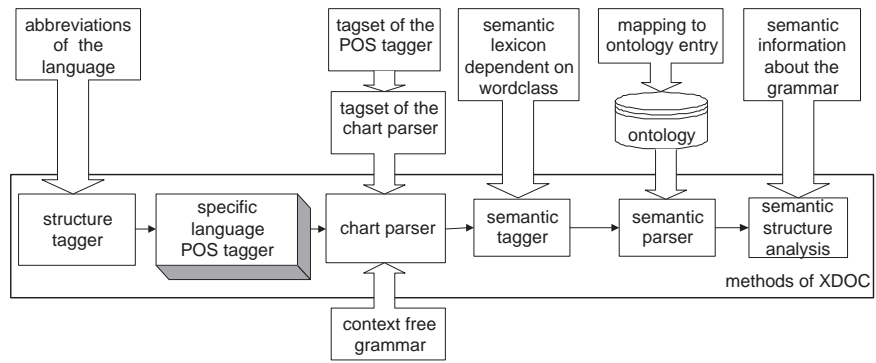

Figure 1: resources of the document suite

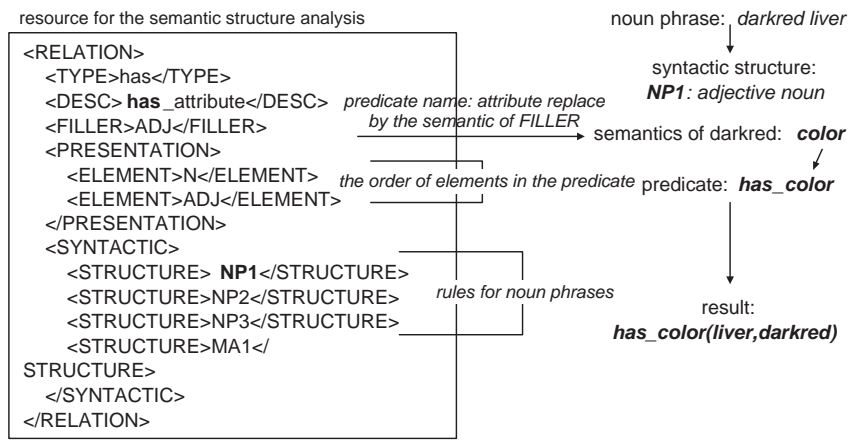

Figure 2: mapping of syntactic structure of noun phrases to the semantic relation 'has'

By resource sharing we can realize multilingual NLP approach with our document suite.